\documentclass{ws-ijns}
\usepackage[utf8]{inputenc}
\usepackage{amssymb}
\usepackage{hhline}
\usepackage{graphicx}
\usepackage{multirow}
\usepackage{url}
\usepackage{ws-rotating}
\usepackage{PTusefulMacros}
\makeatletter
\@namedef{ver@amsmath.sty}{}
\makeatother
\usepackage{amstext}
\makeatletter 
\gdef\@ptsize{0}% 10pt documents 
\let\@currsize\normalsize 
\makeatother 
\usepackage{setspace} 
\def\ptFiguresDirectory#1{./figures/#1}
\def\ivBrack#1{[\![#1]\!]}
\newsavebox\CBox
\def\textBF#1{\sbox\CBox{#1}\resizebox{\wd\CBox}{\ht\CBox}{\textbf{#1}}}
\begin{document}

\markboth{Pawel Trajdos, Marek Kurzynski}
{A Correction Method of a Binary Classifier Applied to Multi-label Pairwise Models.}

\catchline{0}{0}{0000}{}{}

\title{A Correction Method of a Binary Classifier Applied to Multi-label Pairwise Models.}

\author{\footnotesize PAWEL TRAJDOS\footnote{pawel.trajdos@pwr.edu.pl}}

\address{Department of Systems and Computer Networks,\\ Wroclaw
 University of Science and Technology, \\ Wybrzeze Wyspianskiego 27, 50-370
 Wroclaw, Poland
}

\author{MAREK KURZYNSKI\footnote{marek.kurzynski@pwr.edu.pl}}

\address{Department of Systems and Computer Networks,\\ Wroclaw
 University of Science and Technology, \\ Wybrzeze Wyspianskiego 27, 50-370
 Wroclaw, Poland
}

\maketitle

\begin{abstract}
In this work we addressed the issue of applying a stochastic classifier and a local, fuzzy confusion matrix under the framework of multi-label classification. We proposed a novel solution to the problem of correcting label pairwise ensembles. The main step of the correction procedure is to  compute  classifier-specific competence and cross-competence measures, which estimates error pattern of the underlying classifier. We considered two improvements of method of obtaining confusion matrices. The first one is aimed dealing with imbalanced labels. The other, utilizes double labeled instances which are usually removed during the pairwise transformation. 

The proposed methods were evaluated using 29 benchmark datasets. In order to assess the efficiency of the introduced models, they were compared against 1 state-of-the-art approach and the correction scheme based on the original method of confusion matrix estimation. The comparison was performed using four different multi-label evaluation measures: macro and micro-averaged $F_1$ loss,  zero-one loss and Hamming loss. Additionally, we investigated relations between classification quality, which is expressed in terms of different quality criteria, and characteristics of multi-label datasets such as average imbalance ratio or label density.

The experimental study reveals that the correction approaches significantly outperforms the reference method only in terms of zero-one loss. 

\end{abstract}

%%review
\begin{multicols}{2}

\section{Introduction}
\label{sect:Intro}
In many real-world recognition task, there emerges a situation when an object is simultaneously assigned to multiple categories. A great example of such data are tagged photos. An image may be described using such tags as sea, beach and sunset. This kind of dataset is called multi-label data~\cite{Tsoumakas2009} which is becoming more and more common type of data generated by computer systems, and Internet users~\cite{Gantz2012}. Unfortunately, traditional single-label classification methods cannot directly be employed to solve this issue. The main obstacle is that those classifiers are built under assumption that labels do not overlap. In other words, they are capable of predicting only a single category per object which, in case of multi-labelled data, is often insufficient. A solution to this issue is to provide multi-label classification procedures that do not assume that labels are disjoint.

Nowadays, multi-label procedures are utilized in a variety of practical applications but its most widespread applications are: text classification~\cite{Jiang2012} and multimedia classification including classification of video~\cite{Dimou2009}, images~\cite{Xu2011} or music~\cite{Sanden2011}. Another important field of science that extensively uses multi-label models is bioinformatics where multi-label classification is a useful tool for prediction of: gene functions~\cite{Schietgat2010}, protein functions~\cite{Wu2014} or drug resistance~\cite{Heider2013}, to name only a few. 

Our study explores the application of \textit{Random Reference Classifier} (RRC) and local fuzzy confusion matrix to improve the classification quality of the multi-label ensembles using \textit{label-pairwise} decomposition (LPW). The procedure computes label specific competence and cross-competence measures which are used to correct predictions of the classifiers constituting the LPW ensemble. The outcome of each of LPW members is individually modified according to the confusion pattern obtained during the validation stage. And then, they are combined using standard combination methods designed for LPW ensembles. Additionally, we propose two new methods of computing a fuzzy confusion matrix. The first one deals with label imbalance problem via instance weighting. The other incorporates information extracted from instances relevant to multiple labels.

This paper is organized as follows. The next section (Section~\ref{sect:RelWrk}) shows the work related to the issue which is considered in this paper.  The subsequent section (Section~\ref{sect:Proposed}) provide a formal notation used throughout this article, and introduces the proposed algorithm. Section~\ref{sect:ExpSet} contains a description of experimental setup. In section~\ref{sect:ResAndDisc} the experimental results are presented and discussed. Finally, section~\ref{sect:Conc} concludes the paper. 

\section{Related Work}
\label{sect:RelWrk}
Multi-label classification algorithms can be broadly divided into two main groups: set transformation algorithms and algorithm adaptation approaches~\cite{Tsoumakas2009,Gibaja2014}. Algorithm adaptation methods are based upon existing multi-class methods which are tailored to solve multi-label classification problem directly. A great example of such methods are Multi-label back propagation method for artificial neuron networks~\cite{MinLingZhang2006}, multi label KNN algorithm~\cite{Zhang2007}, the ML Hoeffding trees~\cite{Read2012}, the Structured SVM approach~\cite{Diez2014}. Another branch of research that falls under the algorithm adaptation framework is to adapt known deep learning algorithms to solve multi-label task~\cite{Wei2015}.

On the other hand, methods from the former group transform original multi-label problem into a set of single-label classification problems and then combine their output into multi-label prediction. The simplest and most intuitive method from this group is the \textit{Binary Relevance} approach (known also as one-vs-rest approach) that decomposes multi-label classification into a set of binary classification problems. The method assigns an one-vs-rest classifier to each label. Although this method offers a great scalability, which is a desired property in domains with high number of labels~\cite{Tsoumakas_Katakis_Vlahavas_2008}, it also makes an unrealistic assumption that labels are independent. As a consequence, the approach offers acceptable classification quality, however it can easily be outperformed by algorithms that considers dependencies between labels~\cite{Read2009,AlvaresCherman2010}. 

The second approach is \textit{label powerset} method that encodes each combination of labels into a separate meta-class~\cite{Tsoumakas2011}. Although, this transformation allows to incorporate all possible between-label dependencies, it also causes exponential growth of model complexity. As a consequence, the method can be successfully applied when the number of labels is small. Otherwise, the computational burden becomes unacceptable. What is more, when the number of classes is large, the classifier tends to be cursed by overfitting. Nevertheless, this issue was tackled successfully using ensemble systems~\cite{Read2008,Tsoumakas2011}.

Another technique of decomposition of multi-label classification task into a set of binary classifiers is the \textit{label-pairwise} scheme. Under this framework, a binary classifier is trained for each pair of labels, and its task is to separate given labels. The outcome of the classifier is interpreted as an expression of pairwise preference in a label ranking~\cite{Hllermeier2010}. In other words, the classification outcome shows which label is preferred within the pair. Finally, outputs of binary models are collected and a final ranking is formed using chosen merging procedure~\cite{Hllermeier2008}. To convert the ranking into a binary response a thresholding procedure must be employed~\cite{Hllermeier2010}. 

This method of decomposition results in higher computational burden in comparison to the BR-based procedures.On the other hand, pairwise decision boundaries can be considerable simpler to estimate than those originated from BR transformation. As a consequence, the induced models forming the LPW ensemble are less complex that corresponding models of the BR ensemble~\cite{Furnkranz2002}.

One of the main drawbacks of LPW systems is that the direct application of paired comparisons does not perform well on the multi-label datasets in which labels significantly overlaps~\cite{Petrovskiy2006}. Petrovskiy~\cite{Petrovskiy2006} tackled this problem by introducing a pair of binary classifiers for each pair of labels. Within each pair, the classifiers collaborate in order to detect label specific areas and overlapping area. Strictly speaking, each classifier is designed to separate target class combined with overlapping area against the other label.  Finally, the results are combined using Bradley-Terry model. Similar approach was proposed by Wang et al.~\cite{ShuPengWan2007}. They build two classifiers per label pair. The first one is trained to recognize points relevant to both labels. The other is an ordinary pairwise classifier. During the inference phase, the classifiers are pooled sequentially. If the first model marks an example as multi-labelled, the second model is not queried. An alternative procedure is proposed by Chen et al~\cite{Chen2010} who proposed to combine  LPW ensemble with BR ensemble. They employed probabilistic SVM models combined with \textit{delicate boundary} approach to predict four possible classes in the pairwise problem~\cite{Chen2010b}. The label overlapping problem can also be solved using SVM models with two parallel separating hyperplanes~\cite{wang_chang_feng_2005}.

Although the computational complexity of training phase of LPW cannot be reduced, it is possible to trim the inference step. The simplest idea to reduce the computational burden of LPW ensemble is to build an effective binary classifier that allows the ensemble to scale well with the number of instances and the number of labels. This possibility was explored by Menc\'{i}a and F\"{u}rnkranz. They proposed an effective perceptron classifier that can successfully be applied to any label transformation approach including label-pairwise methods~\cite{LozaMenca2008}.
One of another possible approaches is the Quick Weighted Voting (QWeight) method~\cite{LozaMenca2010}. The procedure incorporates the idea of insertion an artificial calibration label~\cite{Frnkranz2008}. This technique offers an alternative to learning additional model that predicts threshold which separates relevant and irrelevant labels. Namely, all relevant labels should obtain more votes than the calibration label, and irrelevant should get fewer votes. Votes for the calibration label are obtained using an additional BR classification system and all votes against real labels are treated as votes for calibration label. QWeighted avoids pooling classifiers whose vote can not change the final outcome of the classification process. As a consequence, the number of queried classifiers is significantly reduced.
Inspired by CLR approach~\cite{Frnkranz2008}. Madjarov et al proposed \textit{two stage voting method} (TSVM) which also combines BR and LPW ensembles~\cite{Madjarov2012}. The learning phase of the method is identical to the learning procedure of CLR. During the inference phase, on the other hand,  the BR ensemble  acts as a filter that decides if second layer classifiers are queried. 

The concept of the \textit{fuzzy confusion matrix} (FCM) was first introduced in studies related to the task of hand gestures recognition~\cite{Kurzynski2015}. The study was focused on multi-class problem with hierarchical structure of gesture classes. The proposed system uses two main advantages of FCM approach. That is, its ability to correct output of a classifier that makes systematic errors. The other is a possibility of handling imprecise class assignment. Experimental study showed that the model offers classification quality comparable to the reference models. What is more important, it is able to utilize imprecise information in a more effective way. The model was also evaluated using a larger number of multi-class datasets~\cite{Trajdos2016}.

Above-mentioned approach was employed under multi-label classification framework~\cite{Trajdos2015,Trajdos2015b}. Namely, it was used to improve the quality of binary relevance classifiers. Experiments confirmed the validity of its use, but also showed sensitivity to the unbalanced class distribution in a binary problem. In this study, we are focused on addressing this issue via employing other technique of multi-label decomposition.

\section{Proposed method}
\label{sect:Proposed}

\subsection{Preliminaries}
\label{subsect:Prelim}
As it was highlighted in the introductory section, multi-label (ML) classification is a generalization of single-label classification task. Under the ML formalism a $d-\mathrm{dimenstional}$ object $\vec{x}=\ptVec2{x}{}{d}\in\mathcal{X}$ is assigned to a set of labels indicated by a binary vector of length $L$: $\vec{y}=\ptVec2{y}{}{L}\in\mathcal{Y}=\{0,1\}^{L}$, where $L$ denotes the number of labels. Each element of the vector is related to a single label and $y_{i}=1$ ($y_{i}=0$) denotes that $i-\mathrm{th}$ label is relevant (irrelevant) for the object $\vec{x}$. 
In this study we suppose, that  multi-label classifier $\psi$, which  maps feature space $\mathcal{X}$ to the set $\mathcal{Y}$, is built in a supervised learning procedure using the training set $\mathcal{S}_N$ containing $N$ pairs of feature vectors $\vec{x}$ and corresponding class labels $\vec{y}$.

Additionally, throughout this paper we follow the statistical classification framework, so vectors $\vec{x}$ and $\vec{y}$ are treated as realisations of random vectors ${\vec{\textbf{X}}=\ptVecA{\textbf{X}}{d} }$ and ${\vec{\textbf{Y}}=\ptVecA{\textbf{Y}}{L}}$, respectively.

Staying with the statistical classification, let us define the Bayesian Classifier for the ML recognition problem. For this purpose, we made an assumption that the joint distribution $P(\vec{\textbf{X}},\vec{\textbf{Y}})$ on $\mathcal{X}\times\mathcal{Y}$ is known. However, in actual classification problems aforesaid assumption does not hold, and the distribution must be estimated on the basis of a training set $\mathcal{S}_N$. Additionally, we define a loss function $\mathcal{L}: \mathcal{Y}\times\mathcal{Y}\mapsto\mathcal{R}_{+}$ that assess the similarity of output space vectors. Considering all the above, the Bayesian Classifier is defined as a classifier $\psi^{*}$  that minimizes the expected loss over the joint probability distribution $P(\vec{\textbf{X}},\vec{\textbf{Y}})$ or equivalently as the point-wise optimal decision for instance $\vec{x}$:
\begin{eqnarray}\label{eq:optBayesClass}
 \psi^{*}(\vec{x}) &=& \argmin_{\vec{h}\in\mathcal{Y}}\sum_{\vec{y}\in\mathcal{Y}}\mathcal{L}(\vec{h},\vec{y})P(\vec{y}|\vec{x})
\end{eqnarray}
where $P(\vec{y}|\vec{x})=P(\vec{\textbf{Y}}=\vec{y}|\vec{\textbf{X}}=\vec{x})$ is the conditional probability of vector $\vec{y}$ given an object $\vec{x}$.

\subsection{Pairwise Transformation}
\label{subsect:PWT}

As it was mentioned in the introductory section, in the label-pairwise (LPW) transformation, also known as \textit{One vs One} (OvO) transformation, the multi-label classifier $\psi$ is decomposed into an ensemble of binary classifiers (called LPW ensemble) $\Psi$ for each pair of labels:
\begin {equation}   
\Psi=\{\psi_{m},\; m=1,2,...,L(L-1)/2\}.
\end{equation}
At the continuous-valued output level, classifier $\psi_{m}(\vec{x})$ produces a 2-dimensional vector of label supports $\left[d^{m1}_{m}(\vec{x}), d^{m2}_{m}(\vec{x})\right] \in [0,1]^{2}$, which values are interpreted as the supports for the hypothesis that  $m_{1}$--th and $m_{2}$--th labels are relevant for the object $\vec{x}$. Without loss of generality we assume that $d^{m_1}_{m}(\vec{x}) + d^{m_2}_{m}(\vec{x})=1$. 

The label-valued outcome can be received according to the maximum rule:
\begin{equation}
\psi_m(\vec{x})=h_m  \longleftrightarrow    d^{h_m}_m(\vec{x}) =\max \{d^{m_{1}}_{m}(\vec{x}), d^{m_{2}}_{m}(\vec{x})\}.
\end{equation}

All binary classifiers in the LPW ensemble $\Psi$ contribute to the final decision through combining their continuous-valued or label-values outputs.  In the first approach the normalized total support for $i$-th label is calculated:
\begin{equation}\label{eq:softRank}
d^{(i)}(\vec{x})= \frac{1}{L-1}\sum_{m: m_k = i}  d_m^{m_k}(\vec{x}), \; i=1,2,...,L,
\end{equation}
whereas in the later case votes for $i$-th label are counted and also normalized ($\ivBrack{.}$ denotes the Iverson bracket):
\begin{equation}\label{eq:crispRank}
n^{(i)}(\vec{x}) = \frac{1}{L-1}\sum_{m} \ivBrack{\psi_m(\vec{x})=i},  \; i=1,2,...,L.
\end{equation}
Final multi-label classification, i.e. response of multi-label classifier $\psi(\vec{x})$:
\begin{equation}  
\psi(\vec{x})=[h^{(1)}(\vec{x}), h^{(2)}(\vec{x}),\ldots,h^{(L)}(\vec{x})],  
\end{equation}
is obtained as a result of thresholding procedure, i.e.
\begin{equation}\label{eq:rnkThr1}  
h^{(i)}(\vec{x})=\ivBrack{d^{(i)}>\theta}
\end{equation}
or
\begin{equation}\label{eq:rnkThr2}  
h^{(i)}(\vec{x})=\ivBrack{n^{(i)}>\theta}
\end{equation}
where acceptance threshold $\theta$ is usually set to $0.5$. 

In the training of  binary classifier $\psi_m$ only learning objects belonging  either to $m_1$-th or $m_2$-th label are used. Examples common to both labels, and remaining labels are ignored~\cite{Hllermeier2010}.

\subsection{Proposed Correction Method}
\label{subsect:COR}

The proposed correction method is based on an assessment of the probability of classifying an object $\vec{x}$ to the class $h_m \in \{m_1, m_2\}$  by the binary classifier $\psi_m$. 
Such an approach requires a probabilistic model which assumes that result of classification $h_m$ of object $\vec{x}$ by binary classifier $\psi_m$, true label $s_m \in \{m_1, m_2\}$ and feature vector $\vec{x}$ are observed values of random variables \textbf{$H_m$}$(\vec{x})$, \textbf{$S_m$}, $\vec{\textbf{X}}$, respectively. 
Random $s_m$ and $\vec{x}$ is a simple consequence of the probabilistic model presented in the previous subsection. 

Random $h_m$ for a given $\vec{x}$ denotes that binary classifier $\psi_m$ is a randomized classifier which  is defined by the conditional probabilities $P(\psi_m(\vec{x})=h_m)=P(h_m|\vec{x}) \in [0,1]$~\cite{Berger1985}. For deterministic classifier these probabilities are equal  $0$ or $1$. 

Injection of the additional random variable into the Bayesian model allows us to define the posterior probability  $P(s_m|\vec{x})$ of label $s_m$ as:
\begin{eqnarray}\label{eq:postProb1}
 P(s_m|\vec{x}) &=& \sum_{h_m \in \{m_1, m_2\}} P(h_m|\vec{x}) P(s_m|h_m,\vec{x}). 
\end{eqnarray}
where $P(s_m|h_m,\vec{x})$ denotes probability that an object $\vec{x}$ belongs to the class $s_m$ given that $\psi_m(\vec{x})=h_m$. 

Probabilities in (\ref{eq:postProb1}) can be evaluated in the terms of competence of binary classifier $\psi_m$  understood as its capability  to the correct activity~\cite{kun2004}. 
%%%%
The competence index is proportional to the probability of correct classification $c_{s_m,s_m}^{(m)}(\vec{x})\approx P(s_m|s_m,\vec{x})$, whereas the cross-competence follows the probability of miss-classification $c_{s_m,h_m}^{(m)}(\vec{x})\approx P(s_m|h_m,\vec{x}), s_m\neq h_m$.

Unfortunately, at the core of the proposed method, we put rather an impractical assumption that the classifier assigns a label in a stochastic manner. We dealt with this issue by harnessing deterministic binary classifiers whose statistical properties were modelled using the RRC procedure~\cite{Woloszynski2011}. The RRC model calculates the probability that the underlying classifier assigns an instance to class $h_m$: $P(h_m|\vec{x})\approx P^{(RRC)}(h_m|\vec{x})$. As a consequence, the final prediction is:
\begin{equation}\label{eq:postProbF}
 P(s_m|\vec{x})\approx\sum_{h_m \in \{m_1,m_2 \}}^{1}P^{(RRC)}_{m}(h_m|\vec{x})c_{s_m,h_m}^{(m)}. 
\end{equation}

%%dodać opis RRC
\subsection{Randomized Reference Classifier}
\label{subsect:RRC}

The behaviour of a base classifier $\psi_{m}$ is modelled using a stochastic classifier defined using a probability distribution over the set of labels $\{m_1,m_2 \}$. In this work employed the randomized reference classifier (RRC) proposed by Woloszynski and Kurzynski~\cite{Woloszynski2011}.For given instance $\vec{x}$, the RRC $\psi_{m}^{(RRC)} $ generates a vector of class supports  $\left[ \delta^{m_1}_{m}(\vec{x}), \delta^{m_2}_{m}(\vec{x})\right]$ which are observed realisations underlying of random variables $\left[ \Delta^{m_1}_{m}(\vec{x}), \Delta^{m_2}_{m}(\vec{x})\right]$. The support vector is converted into classifier decision using maximum rule. Probability distribution of random variables is chosen in such a way that the following conditions are satisfied:
\begin{eqnarray}
\label{MK_PT:delta_c1}
\Delta_{m_1}(\vec{x}), \, \Delta_{m_2}(\vec{x}) \in (0,1)\\
\label{MK_PT:delta_c2}
\Delta_{m_1}(\vec{x})+\Delta_{m_2}(\vec{x})=1\\
\label{MK_PT:delta_c3}
\mathrm{E}\left[\Delta^{m_i}_{m}(\vec{x}) \right] = d^{m_i}_{m}(x),\ i \in \{1,2 \},
\end{eqnarray}
where $\mathrm{E}$ is the expected value operator. Conditions (\ref{MK_PT:delta_c1})  and (\ref{MK_PT:delta_c2})  follow from the normalisation properties of class supports while condition (\ref{MK_PT:delta_c3}) ensures the equivalence of the randomized model $\psi_l^{(RRC)}$ and underlying base classifier $\psi_{m}$.

Since the distribution of random variables is known, it is possible to calculate the probability of classification an object $\vec{x}$ to the $h_m$-th class:
\begin{equation}   \label{MK_PT:wzor5}
P^{(RRC)}(h_m|x)=Pr\left[\Delta^{h_m}_{m}(\vec{x})> \Delta^{\{m_1,m_2 \} \setminus h_m}_{m}(\vec{x})\right].
\end{equation}

During the designing process of RRC, the cruical step is to chose the probability distributions for random variables $\Delta^{m_i}_{m}(\vec{x})\;  i \in \{1,2\}$ so that the conditions~(\ref{MK_PT:delta_c1})-(\ref{MK_PT:delta_c3}) are met. In this study, we follow the recomendations made by the authors of RRC and we employed beta distributions with parameters  $\lambda^{m_i}_{m}(\vec{x}),\mu^{m_i}_{m}(\vec{x}),\;  i \in \{1,2\}$. The justification of the choice of the above-mentioned distribution and detailed description of $\lambda,\:\mu$ parameters estimation can be found  in \cite{Woloszynski2011}.

For beta distribution we get the following formula for probability (\ref{MK_PT:wzor5}):

\begin{equation}  \nonumber
P^{(RRC)}(h_m|\vec{x})  = \int_0^1 b(u,\lambda^{m_i}_{m}(\vec{x}), \mu^{m_i}_{m}(\vec{x})) \times
\end{equation}
\begin{equation} \label{MK_PT:wzor6}  
\times \left[\prod_{i \in \{m_1, m_2\} } B(u,\lambda^{m_i}_{m}(\vec{x}), \mu^{m_i}_{m}(\vec{x})\right]\ du,
\end{equation}

where $B(\dot)$ is a beta cumulative distribution function and $b()$ is a beta probability density function.

\subsection{Confusion Matrix}
\label{subsect:ConfM}

During the inference process of the proposed approach, we must estimate the probability $P(s_m|h_m,\vec{x})$. To achieve this goal we created an estimation procedure based on local, fuzzy confusion matrix. A confusion matrix provides us with a complete information on classification accuracy for separate classes~\cite{Sindhwani2001}. An example of such matrix for a binary classification task is given in Table~\ref{MK_PT:confmatrix}. The rows of the matrix corresponds to the ground-truth classes, whereas the columns match the outcome of a classifier. The fuzzy nature of the confusion matrix arises directly from the fact that a stochastic model has been employed. In contrast to the decision regions produced by a deterministic classifier, decision regions of the random classifier must be expressed in terms of fuzzy set formalism~\cite{Zadeh1965}. In order to provide an accurate estimation, we have also defined our confusion matrix as local which means that the matrix is build using neighbouring points of the instance $\vec{x}$. Under assumption that a prior distribution of the label $P(s_m)$, class-conditional distribution of features $P(\vec{x}|s_m)$ and stochastic classifier distribution  $\Psi(\vec{x})$ are known, each element of the confusion matrix is calculated by performing the following integration:
\begin{equation}
\label{MK_PT:cmfc}
\varepsilon_{s_m,h_m}^{m}(\vec{z})= P(s_m)\int_{\mathcal{X}}P(\vec{x}|s_m)\mu_{\mathcal{D}^{m}_{h_m}}(\vec{x})\mu_{\mathcal{N}(\vec{z})}(\vec{x})\,d\vec{x},
\end{equation}%
where $\vec{z}\in\mathcal{X}$ is an instance around which we construct the confusion matrix, and $\mu_{\mathcal{D}^{m}_{h_m}}(x)=P^{(RRC)}(h_m|\vec{x})$ indicates the fuzzy decision region of the stochastic classifier. Additionally, $\mu_{\mathcal{N}(\vec{z})}(\vec{x})$ denotes the fuzzy neighbourhood of instance $\vec{z}$. The membership function of the neighbourhood was defined using Gaussian potential function:
\begin{equation}   \label{MK_PT:mu}
 \mu_{\mathcal{N}(\vec{x})}(\vec{z})=\exp({-\beta \delta(\vec{z},\vec{x})^2}), 
\end{equation}
where $\beta \in \mathbb{R}^{+}$ and $\delta(\vec{z},\vec{x})$ is the Euclidean distance between $\vec{z}$ and $\vec{x}$. During the experimental study, $\beta$ was set to $1$.

%%%%%%%%%FCM
\begin{tablehere}
\vbox{%
\centering
\tbl{The confusion matrix for a binary classification problem.\label{MK_PT:confmatrix}}
{\begin{tabular}{@{}cc|cc@{}}
& & \multicolumn{2}{c}{estimated}\\
& &  $h_m=m_1$ & $h_m=m_2$\\
\hline
\multirow{2}{*}{true}& $s_m=m_1$& $\varepsilon_{m_1,m_1}^{m}$&$\varepsilon_{m_1,m_2}^{m}$\\
& $s_m=m_2$& $\varepsilon_{m_2,m_1}^{m}$&$\varepsilon_{m_2,m_2}^{m}$\\
\end{tabular}
}
}
\end{tablehere}

\subsubsection{Confusion matrix for balanced data}
\label{subsubsect:confMBal}

Now, let's tackle a more practical case. In real-world classification tasks, the assumption about acquaintance of the distributions $P(s_m)$, $P(\vec{x}|s_m)$ and $\Psi_{m}(\vec{x})$  does not hold, while a validation set is available:
\begin{equation}\label{eq:valSet} 
\mathcal{V}=\left\{(\vec{x}^{(1)},\vec{y}^{(1)}), (\vec{x}^{(2)},\vec{y}^{(2)}), \ldots ,(\vec{x}^{(M)},\vec{y}^{(M)})\right\},
\end{equation}
where $\vec{x}^{(k)} \in \mathcal{X},\ \vec{y}^{(k)} \in \mathcal{Y}$. 
On the basis of this set we define pairwise subsets of validation set, fuzzy decision region of $\psi_{m}$ and set of neighbours of $\vec{z}$:
\begin{eqnarray}\label{eq:fvs}
\mathcal{V}^{m}_{s_m} &=& \left\{  (\vec{x}^{(k)},\vec{y}^{(k)}, 1): (\vec{x}^{(k)},\vec{y}^{(k)}) \in \mathcal{V},\right.\nonumber \\
  & &\left. y^{(k)}_{m_1} + y^{(k)}_{m_2}=1,y^{(k)}_{s_m}=1  \right\},
\end{eqnarray}
\begin{equation}\label{eq:fds}
 {\mathcal{D}}^{m}_{h_m} = \left\{  (\vec{x}^{(k)},\vec{y}^{(k)} , \mu_{\mathcal{D}^{m}_{h_m}}(\vec{x}^{(k)}  ) ): (\vec{x}^{(k)},\vec{y}^{(k)}) \in \mathcal{V}\right\},
\end{equation}
\begin{equation}\label{eq:fns}
 \mathcal{N}(\vec{z}) = \left\{  (\vec{x^{(k)}},\vec{y}^{(k)} ,\mu_{\mathcal{N}(\vec{z})}(\vec{x^{(k)}}) ):(\vec{x}^{(k)},\vec{y}^{(k)}) \in \mathcal{V} \right\},
\end{equation}
where each triplet $(\vec{x}^{(k)},\vec{y}^{(k)}, \zeta)$ defines fuzzy membership value $\zeta$ of instance $(\vec{x}^{(k)},\vec{y}^{(k)})$.
The following fuzzy sets are employed to approximate entries of the local confusion matrix (\ref{MK_PT:cmfc}):
\begin{eqnarray}\label{eq:uFCM}
\hat{\varepsilon}^{m}_{s_m,h_m}(\vec{z}) &=& \frac{|\mathcal{V}^{m}_{s_m} \cap {\mathcal{D}}^{m}_{h_m} \cap \mathcal{N}(\vec{z})|}{|\mathcal{N}(\vec{z})|}
\end{eqnarray}
where $|.|$ is the cardinality of a fuzzy set~\cite{Dhar2013}. Finally, the approximation of $P(s_m|h_m,\vec{x})$ is calculated as follows:
\begin{equation} \label{pt:postApprox}
c_{s_m,h_m}^{m}(\vec{z})  = \frac{\hat{\varepsilon}^{m}_{s_m,h_m}(\vec{z})}{\sum_{u \in \{ m_1,m_2 \}}\hat{\varepsilon}_{u,h_m}^{m}(\vec{z})}.
\end{equation}

\subsubsection{Confusion matrix for imbalanced data}
\label{subsubsect:CSConfM}

As previously mentioned, the proposed method is sensitive to class imbalance ratio. However, we employed LPW transformation, which is known to be less imbalanced than BR transformation method. However some of the transformed problems still remain imbalanced (see Section~\ref{sect:ExpSet}, Table~\ref{table:Dataset_summ}). To deal with this issue, we should reduce the impact of majority class on the estimation of the confusion matrix. We achieve this by assigning each instance a class-specific weight which is defined as follows:
\begin{equation}\label{eq:weightFac}
 w_{s_m}^{m} = \min(1, \frac{\left| \mathcal{V}^{m}_{\{m_1,m_2 \}\setminus s_m}\right| }{\left| \mathcal{V}^{m}_{s_m} \right| }).
\end{equation}
The weighting scheme guarantees that:
\begin{equation}
 w_{s_m}^{m}\left| \mathcal{V}^{m}_{s_m} \right| = w_{\{m_1,m_2 \}\setminus s_m}^{m} \left| \mathcal{V}^{m}_{\{m_1,m_2 \}\setminus s_m} \right|
\end{equation}

To incorporate instance weighting into confusion matrix estimation we defined a new fuzzy set:
\begin{eqnarray}\label{eq:fws}
\mathcal{W}^{m}_{s_m} &=& \left\{  (\vec{x}^{(k)},\vec{y}^{(k)},w_{s_m}^{m} ):(\vec{x}^{(k)},\vec{y}^{(k)}) \in \mathcal{V}\right. \nonumber \\
  &,&\left. y^{(k)}_{m_1} + y^{(k)}_{m_2}=1,y^{(k)}_{s_m}=1  \right\},
\end{eqnarray}
which extends the estimation formula:
\begin{equation}\label{eq:wFCM}
\hat{\varepsilon}^{m}_{s_m,h_m}(\vec{z}) = \frac{|\mathcal{V}^{m}_{s_m} \cap {\mathcal{D}}^{m}_{h_m} \cap \mathcal{N}(\vec{z}) \cap \mathcal{W}^{m}_{s_m}|}{|\mathcal{N}(\vec{z}) \cap \mathcal{W}^{m}_{s_m}|}
\end{equation}

\subsubsection{Confusion matrix -- overlaping labels}
\label{subsubsect:OLConfM}

The original method of LPW decomposition requires training each binary classifier using a training subspace in which instances shared between both labels are removed~\cite{Hllermeier2010}. However, during the validation phase, the inclusion of these instances may improve estimation of classifier competence in regions of the input space dominated by points relevant to both labels. To include multi-labeled instances we extend set $\mathcal{V}_{s_m}^{m}$ according to the following equation:
\begin{eqnarray}\label{eq:fvso}
\widetilde{\mathcal{V}}^{m}_{s_m} &=& \mathcal{V}^{m}_{s_m} \cup \left\{  (\vec{x}^{(k)},\vec{y}^{(k)}, 0.5):\right. \nonumber \\
 & & \left.(\vec{x}^{(k)},\vec{y}^{(k)}) \in \mathcal{V}, y^{(k)}_{m_1} + y^{(k)}_{m_2}=2\right\}.
\end{eqnarray}
The extended set replaces the original one during estimation of confusion matrix entries.

\subsection{System Architecture}
\label{subsect:SystemArch}

The description of learning and inference phases are provided in Figures~\ref{pt:table:Train} and~\ref{pt:table:Test}.
\begin{figurehere}
\vbox{%
\begin{center}
\caption{Pseudocode of the learning procedure.}
\label{pt:table:Train}%
\hrule width 0.45\textwidth
\tt
\smallskip
\begin{tabbing}
\quad \=\quad \=\quad \=\quad \=\quad \=\quad \kill
\textbf{Input data:}\\
\>$\mathcal{S}_{N}$ - the initial training set;\\
\>$t \in (0,1)$ -the split percentage;\\
\textbf{BEGIN}\\
\>Split randomly $\mathcal{S}_N$ into $\mathcal{T}$ and $\mathcal{V}$ using $t$:\\
\>\>$|\mathcal{T}| = t|\mathcal{S}_N|$\\
\>\>$|\mathcal{V}| = (1-t)|\mathcal{S}_N|$\\
\>Build the LPW ensemble:\\
\>\>$\Psi=\left\{\psi_{m},\; m=1,2,...,L(L-1)/2\right\}$ using $\mathcal{T}$;\\ 
\>For $1\leq m \leq L(L-1)/2$ and $h_m \in \left\{m_1, m_2 \right\}$\\
\>\>build $\mathcal{D}_{h_m}^{m}$ according to (\ref{eq:fds});\\
\>For $1\leq m \leq L(L-1)/2$ and $s_m \in \left\{m_1, m_2 \right\}$\\
\>\>$\mathcal{V}^{m}_{s_m}$ according to (\ref{eq:fvs}) or (\ref{eq:fvso})\\
\>Save $\mathcal{V}$, $\mathcal{D}_{h}^{m,n}$ and $\mathcal{V}^{m}_{s_m}$;\\
\textbf{END}
\end{tabbing}
\hrule width 0.45\textwidth
\end{center}
}
\end{figurehere}

\begin{figurehere}
\vbox{%
\begin{center}
\caption{The classification procedure for given $\vec{x}$.}%
\label{pt:table:Test}%

\hrule width 0.45\textwidth
\tt
\smallskip
\begin{tabbing}
\quad \=\quad \=\quad \=\quad \=\quad \=\quad \kill
\textbf{Input data:}\\
\>$\mathcal{V}$ - the validation set;\\
\>$\mathcal{D}_{h_m}^{m}$- decision regions:\\
\>\>$1\leq m \leq L(L-1)/2$ and $h_m \in \left\{m_1,m_2 \right\}$;\\
\>$\mathcal{V}_{s_m}^{m}$- subsets of validation sets:\\
\>\>$1\leq m \leq L(L-1)/2$ and $s_m \in \left\{m_1,m_2 \right\}$;\\

\>$\vec{x}$ - the testing point;\\
\textbf{BEGIN}\\
\>build $\mathcal{N}(\vec{x})$;\\
\>For $1\leq m \leq L(L-1)/2$:\\
\>\> calculate $\hat{\varepsilon}^{m}_{s_m,h_m}(\vec{x})$ according to (\ref{eq:wFCM})\\
\>\>\>or (\ref{eq:uFCM});\\
\>\>calculate $c_{s_m,h_m}^{m}(\vec{x}) $ according to (\ref{pt:postApprox});\\
\>\>calculate $P(s_m|\vec{x})$ according to (\ref{eq:postProbF});\\
\>Build final ranking according to (\ref{eq:softRank})\\
\>\>or (\ref{eq:crispRank});\\
\>Convert ranking into\\
\>\>response vector $\psi(\vec{x})$ using (\ref{eq:rnkThr1}) or (\ref{eq:rnkThr2})\\
\> Return $\psi(\vec{x})$;\\
\textbf{END}
\end{tabbing}
\hrule width 0.45\textwidth
\end{center}
}
\end{figurehere}

\section{Experimental Setup}
\label{sect:ExpSet}

The conducted experimental study provides an empirical evaluation of the quality of the proposed methods and compares their results against the outcome of the original LPW approach. Namely, we conducted our experiments using the following algorithms:
\begin{enumerate}
 \item Unmodified LPW classifier~\cite{Hllermeier2008},
 \item LPW classifier corrected using confusion matrix specific to balanced label distributions (Section~\ref{subsubsect:confMBal}, the method is also referred as FCM),
 \item LPW classifier corrected using confusion matrix specific to imbalanced label distributions (Section~\ref{subsubsect:CSConfM}, the method is also referred as FCM-W),
 \item LPW classifier corrected using confusion matrix that considers label overlapping (Section~\ref{subsubsect:OLConfM}, the method is also referred as FCM-O). 
\end{enumerate}
In the following sections of this paper, we will refer to the investigated algorithms using above-said numbers.

All base single-label classifiers were implemented using the Na\"{i}ve Bayes classifier~\cite{Hand2001} combined with Random Subspace technique~\cite{TinKamHo1998}. We utilized Na\"{i}ve Bayes implemented in WEKA framework~\cite{Hall2009}. The classifier parameters were set to its defaults. For the Random Subspace we set the number of attributes to the $20\%$ of the original number of attributes, and the number of repetitions was set to $20$. All multi-label algorithms were implemented using MULAN~\cite{Tsoumakas2011_mulan} framework.

The experiments were conducted using 29 multi-label benchmark sets. The main characteristics of the datasets are summarized in Table~\ref{table:Dataset_summ}. The first column of the table contains set names and its source. The second column gives the ordinal numbers assigned to datasets. The numbers are used throughout the following sections to denote given datasets. Next three columns are filled with the number of instances, dimensionality of the input space and the number of labels respectively. Another four columns provide us with measures of multi-label-specific characteristics of given set i.e. cardinality (LC), density (LD)~\cite{Tsoumakas2009}, average pairwise imbalance ratio (avIR) and average scumble (AVGsc)~\cite{Charte2014}. To be more precise, the mentioned measures describe the following properties of the dataset:
\begin{itemize}
 \item[LC]: the average number of labels per instance,
 \item[LD]: the average number of labels per instance divided by the number of labels,
 \item[avIR]: the indicator of imbalance ratio between labels,
 \item[AVGsc]: describes average label concurrence (co-occurrence between rare and frequent labels) within the set.
\end{itemize}

The extraction of training and test datasets was performed using $10-\mathrm{CV}$. The proportion of the training set $\mathcal{T}$ was fixed at $t=60\%$ of the original training set $\mathcal{S}_N$. The validation set $\mathcal{V}$, and the trainig $\mathcal{T}$ set for LPW-based ensembles do not overlap. The reference method does not utilize the validation set so it was trained using the initial training set $\mathcal{S}_N$. Some of the employed sets needed some preprocessing. That is, multi label regression sets (Flare1/2 and Water-quality) were binarised using thresholding procedure. To be more accurate, when the value of output variable,for given object, is greater than zero, the corresponding label is set to be relevant to the object. We also used multi-label multi-instance~\cite{Zhou2007} sets (No.:2,4,5,12,13,18,20,21) which were transformed to single-instance multi-label datasets according to the suggestion made by Zhou et al.~\cite{Zhou2012}. Namely, the Hausdorff~\cite{rockafellar1998} distance was employed to calculate distances between bag of instances. The calculated distances are used to build distance matrix. Then the multi dimensional scaling~\cite{Borg1997} is used to convert the distance matrix into low dimensional representation of multi-instance objects. Two of used datasets are synthetic ones (SimpleHC, SimpleHS) and they were generated using algorithm described by Torres et al.~\cite{Tomas2014}. To reduce the computational burden we use only two subsets for each of IMDB and Tmc2007 sets.

The experimental study covers two groups of experiments. The first one compares the classification quality gained by proposed methods and the reference procedure. The other investigates relations between classification quality and aforementioned indicators of dataset properties (see Table~\ref{table:Dataset_summ}). The relations were examined using Spearman correlation coefficient~\cite{Spearman1904}.

The algorithms were compared in terms of 4 different quality criteria coming from three groups: instance-based, micro-averaged and macro-averaged~\cite{Luaces2012}. We applied two instance-based criterion, namely zero-one loss and Hamming loss~\cite{Dembczynski2012}. In addition we harness micro/macro-averaged $F_1$-measure~\cite{Luaces2012}. Statistical evaluation of the results was performed using the Friedman~\cite{Friedman1940} test followed by the Nemenyi~\cite{demsar2006}
post-hoc procedure. Additionally, we applied the Wilcoxon signed-rank test~\cite{wilcoxon1945} and the family-wise error rates were controlled using the Bergmann-Hommel's procedure~\cite{Garcia2008}. For all statistical tests, the significance level was set to $\alpha=0.1$.

{
\def\arraystretch{0.7}
\begin{table*}
\centering
\tbl{Summarised properties of the datasets employed in the experimental study. $N$ is the number of instances, $d$ is the dimensionality of input space, $L$ denotes the number of labels. $\mathrm{LC}$, $\mathrm{LD}$, $\mathrm{avIR}$, $\mathrm{AVsc}$ are label cardinality, label density,  average imbalance ratio and label scumble respectively. \label{table:Dataset_summ}}{
\begin{tabular}{l|c||cccccccc}
\hline
Name&	Set No.&	N&	d&	L&	CD&	LD&	avIR&	AVsc\\
\hline
Arts~\cite{Tsoumakas2011_mulan}&		1&	7484&	1759&	26&	1.654&	0.064&	94.738&	0.059\\
Azotobacter~\cite{Wu2014}	&		2&	407&	33&	13&	1.469&	0.113&	2.225&	0.010\\
Birds~\cite{Tsoumakas2011_mulan}&		3&	645&	279&	19&	1.014&	0.053&	5.407&	0.033\\
Caenorhabditis~\cite{Wu2014}&			4&	2512&	41&	21&	2.419&	0.115&	2.347&	0.010\\
Drosophila~\cite{Wu2014}&			5&	2605&	42&	22&	2.656&	0.121&	1.744&	0.004\\
Emotions~\cite{Tsoumakas2011_mulan}&		6&	593&	78&	6&	1.868&	0.311&	1.478&	0.011\\
Enron~\cite{Tsoumakas2011_mulan}&		7&	1702&	1054&	53&	3.378&	0.064&	73.953&	0.303\\
Flags~\cite{Tsoumakas2011_mulan}&		8&	194&	50&	7&	3.392&	0.485&	2.255&	0.061\\
Flare1~\cite{Tsoumakas2011_mulan}&		9&	323&	28&	3&	0.232&	0.077&	2.423&	0.005\\
Flare2~\cite{Tsoumakas2011_mulan}&		10&	1066&	30&	3&	0.209&	0.070&	14.152&	0.006\\
Genbase~\cite{Tsoumakas2011_mulan}&		11&	662&	1213&	27&	1.252&	0.046&	37.315&	0.029\\
Geobacter~\cite{Wu2014}&			12&	379&	31&	11&	1.264&	0.115&	2.750&	0.014\\
Haloarcula~\cite{Wu2014}&			13&	304&	33&	13&	1.602&	0.123&	2.419&	0.016\\
Human~\cite{Xu2013}&				14&	3106&	454&	14&	1.185&	0.085&	15.289&	0.020\\
IMDB0~\cite{meka}&				15&	3042&	1029&	28&	1.987&	0.071&	24.611&	0.109\\
IMDB1~\cite{meka}&				16&	3044&	1029&	28&	1.987&	0.071&	24.585&	0.106\\
Medical~\cite{Tsoumakas2011_mulan}&		17&	978&	1494&	45&	1.245&	0.028&	89.501&	0.047\\
MimlImg~\cite{Zhou2007}&			18&	2000&	140&	5&	1.236&	0.247&	1.193&	0.001\\
Plant~\cite{Xu2013}&				19&	978&	452&	12&	1.079&	0.090&	6.690&	0.006\\
Pyrococcus~\cite{Wu2014}&			20&	425&	38&	18&	2.136&	0.119&	2.421&	0.015\\
Saccharomyces~\cite{Wu2014}&			21&	3509&	47&	27&	2.275&	0.084&	2.077&	0.005\\
Scene~\cite{Tsoumakas2011_mulan}&		22&	2407&	300&	6&	1.074&	0.179&	1.254&	0.000\\
SimpleHC~\cite{Tomas2014}&			23&	3000&	40&	10&	1.900&	0.190&	1.138&	0.001\\
SimpleHS~\cite{Tomas2014}&			24&	3000&	40&	10&	2.307&	0.231&	2.622&	0.050\\
Slashdot~\cite{meka}&				25&	3782&	1101&	22&	1.181&	0.054&	17.693&	0.013\\
Tmc2007\_0~\cite{Tsoumakas2011_mulan}&		26&	2857&	522&	22&	2.222&	0.101&	17.153&	0.195\\
Tmc2007\_1~\cite{Tsoumakas2011_mulan}&		27&	2834&	522&	22&	2.242&	0.102&	17.123&	0.191\\
Water-quality~\cite{Tsoumakas2011_mulan}&	28&	1060&	30&	14&	5.073&	0.362&	1.767&	0.037\\
Yeast~\cite{Tsoumakas2011_mulan}&		29&	2417&	117&	14&	4.237&	0.303&	7.197&	0.104\\
\hline
\end{tabular}}
\end{table*}
}

\section{Results and Discussion}
\label{sect:ResAndDisc}

This section provides the results of the conducted experimental study. The following subsections presents outcomes related to the classification quality, and dependencies between classification quality and set-specific measures.

\subsection{Classification quality}
\label{subsect:classQual}

\begin{figure*}
\begin{center}
  \includegraphics[width=0.60\textwidth]{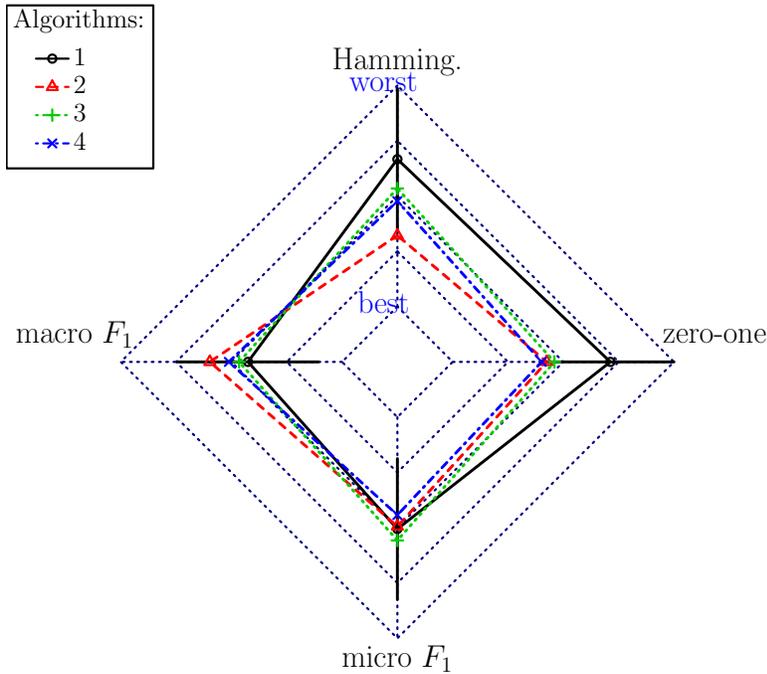}
  \caption{Visualisation of multi-criteria Nemenyi post-hoc test for the investigated algorithms. Algorithms are numbered as in the section describes experimental setup. Black bars parallel to criterion-specific axes denote critical difference for the Nemenyi tests.}%
  \label{figure:radar}
\end{center}
\end{figure*}

The results related to classification quality are presented in tables~\ref{table:Friedman_test}--\ref{table:Full_res} and figure~\ref{figure:radar}. Table~\ref{table:Friedman_test} provides p-values resulting from the application of the Friedman test for each quality criterion. The outcome clearly shows that only for Hamming loss and zero-one loss the differences between algorithms are significant. Those results are also confirmed by the outcome of the Nemenyi post-hoc test. Namely, according to this test, FCM algorithm achieves the classification quality better than unmodified LPW approach under the Hamming loss. What is more, under the zero-one loss, all proposed algorithms significantly improve the reference algorithm. This is an important result because the zero-one loss is known to be the most strict quality criterion that can be employed to assess the outcome of a multi-label classifier. The criterion is also the harshest. That is, it punishes one-label misclassification as hard as misclassification of all labels~\cite{Dembczynski2012}. So significant difference in terms of this measure indicates that the modified algorithms achieved the greatest number of perfect hit results.

Further, the results are also confirmed by the paired Wilcoxon test (tables~\ref{table:PW-Hamming} and~\ref{table:PW-MacroF}). Additionally, the mentioned testing procedure indicates that in most cases there are no significant differences between proposed algorithms.  The only exception is the difference between FCM-O and FCM under the Hamming loss.

The results described above, allow us to make the following conclusions. First of all, FCM is the only algorithm that achieves the significant improvement over the reference method in terms of the Hamming loss. Unfortunately, at the same time, its rank under macro-averaged $F_1$ measure is the greatest. It means that FCM algorithm copes with rare labels in a less efficient way than the other assessed procedures, since micro-averaged $F_1$ measure is more sensitive to quality loss for uncommon labels~\cite{Luaces2012}. The micro-averaged $F_1$ measure in contrast to macro-averaged $F_1$ measure provides better assessment of recognition of frequent labels. As we can see, for micro-averaged $F_measure$ the average ranks obtained by compared algorithms are very close to each other. Moreover, for multi-label datasets characterised by low label-density (LD), it is easy to obtain good classification quality by setting all rare labels as irrelevant. Taking all this into account, we can make a conclusion that FCM algorithm remains sensitive to imbalanced label distributions despite the use of label-pairwise transformation. The approach still produces too much false negative responses, although the imbalance ratio of LPW transformation is lower than imbalance ratio related to BR approach.

The FCM-W algorithm was designed to improve classification quality when label imbalance ratio is high. The experimental results confirmed that it performs better than FCM algorithm. Namely, contrary to FCM approach, its performance under the Hamming loss and macro-averaged $F_1$ loss does not express undesirable behaviour. Furthermore, in terms of remaining criteria, relative rank differences are rather small. 

In addition, it is worth to notice that FCM-O algorithm achieves similar classification quality to FCM-W algorithm. This observation shows that the impact of imbalanced label distribution can also be effectively mitigated via inclusion of double labelled instances. Insertion of additional instances instead of increasing weights of existing ones not only reduce imbalance factor but also reduces the risk of overfitting. From this perspective, FCM-O classifier should be preferred to FCM-W.

%%Friedman
{
\begin{tablehere}
\vbox{%
\centering
\tbl{The outcome (p-value) of the Friedman statistical test conducted for rankings built under different criteria. \label{table:Friedman_test}}{
\begin{tabular}{rr}
  \hline
 Criterion& p-value \\ 
  \hline
  Hamming &	 0.063198 \\ 
  macro $F_1$&	 0.910818 \\ 
  micro $F_1$ &	 0.910818 \\ 
  zero-one&	 0.007656 \\ 
   \hline
\end{tabular}
}
}
\end{tablehere}
}

%%Hamming & macro
{
\begin{table*}
\centering
\tbl{Wilcoxon test -- p-values for paired comparisons of investigated algorithms for Hamming loss and macro $F_1$ loss respectively. Algorithms are numbered according to Section~\ref{sect:ExpSet}. Due to the symmetry of the following table, some entries are omitted. The last row of the table presents average ranks achieved over the test sets. \label{table:PW-Hamming}}{
\begin{tabular}{@{}r|cccc||}
  \hline
  &\multicolumn{4}{c||}{Hamming}\\
  \hline
 & 1 & 2 & 3 & 4 \\ 
  \hline
  1 &  & 0.063 & 0.162 & 0.152 \\ 
  2 &  &  & 0.152 & 0.063 \\ 
  3 &  &  &  & 0.152 \\ 
  \hline
  Rnk & 3.000 & 1.966 & 2.603 & 2.431 \\ 
   \hline
\end{tabular}%
\begin{tabular}{r|cccc@{}}
  \hline
  &\multicolumn{4}{c}{macro $F_1$}\\
  \hline
 & 1 & 2 & 3 & 4 \\ 
\hline
 1 &  & 0.413 & 0.608 & 0.413 \\ 
  2 &  &  & 0.413 & 0.608 \\ 
  3 &  &  &  & 0.608 \\ 
  \hline
Rnk & 2.276 & 2.793 & 2.397 & 2.534 \\ 
   \hline
\end{tabular}
}
\end{table*}
}

%%Micro F & zero-one
{
\begin{table*}
\centering
\tbl{Wilcoxon test -- p-values for paired comparisons of investigated algorithms for micro $F_1$ loss and zero-one loss respectively. \label{table:PW-MacroF}}{
\begin{tabular}{@{}r|cccc||}
  \hline
  &\multicolumn{4}{c||}{micro $F_1$}\\
  \hline
 & 1 & 2 & 3 & 4 \\ 
\hline
  1 &  & 1.000 & 0.413 & 1.000 \\ 
  2 &  &  & 0.413 & 1.000 \\ 
  3 &  &  &  & 0.413 \\ 
  \hline
  Rnk & 2.517 & 2.483 & 2.672 & 2.328 \\ 
   \hline
\end{tabular}%
\begin{tabular}{r|cccc@{}}
  \hline
  &\multicolumn{4}{c}{zero-one}\\
  \hline
 & 1 & 2 & 3 & 4 \\ 
\hline
  1 &  & 0.054 & 0.040 & 0.054 \\ 
  2 &  &  & 1.000 & 1.000 \\ 
  3 &  &  &  & 1.000 \\ 
  \hline
Rnk & 3.138 & 2.276 & 2.379 & 2.207 \\ 
   \hline
\end{tabular}
}
\end{table*}
}

%%rank test -- Nemenyi

%full
{
\begin{sidewaystable*}
\centering
\tbl{Full set-specific results for all considered quality criteria. Algorithms are numbered according to Section~\ref{sect:ExpSet}. Sets are numbered according to Table~\ref{table:Dataset_summ}. Each table entry represents the set-specific loss value, which is averaged over 10 folds, for given algorithm. Entries corresponding to the best criterion value are highlighted using boldface. The last row of the table presents average ranks achieved over the test sets. \label{table:Full_res}}{
\begin{tabular}{@{}|l||cccc|cccc|cccc|cccc|@{}}
\hline
 & \multicolumn{4}{c|}{Hamming} & \multicolumn{4}{c|}{macro $F_1$}& \multicolumn{4}{c|}{micro $F_1$}& \multicolumn{4}{c|}{zero-one}\\
\hline
{\scriptsize Set No.}&1&2&3&4&		1&2&3&4&		1&2&3&4&		1&2&3&4\\
\hline
1&.484&\textBF{.452}&.578&.453&				\textBF{.817}&.824&.846&.825&			.802&\textBF{.794}&.831&\textBF{.794}&		1.00&1.00&1.00&1.00\\
2&.167&.167&\textBF{.157}&.167&				.980&.980&\textBF{.979}&.980&			.881&\textBF{.880}&.890&\textBF{.880}&		.995&.995&\textBF{.922}&.995\\
3&.487&\textBF{.478}&.485&.481&				.852&\textBF{.847}&.849&\textBF{.847}&		.845&\textBF{.838}&.840&\textBF{.838}&		1.00&1.00&1.00&1.00\\
4&.143&\textBF{.136}&\textBF{.136}&\textBF{.136}&	\textBF{.991}&.993&.993&.993&			\textBF{.938}&.954&.954&.954&		.895&\textBF{.791}&\textBF{.791}&\textBF{.791}\\
5&.149&.137&.137&\textBF{.136}&				\textBF{.990}&.994&.994&.994&			\textBF{.930}&.961&.961&.961&		.943&.744&.744&\textBF{.741}\\
6&.308&\textBF{.306}&.313&.311&				.368&.362&\textBF{.358}&.360&			.366&.364&.364&\textBF{.363}&		\textBF{.895}&.902&.923&.907\\
7&.399&\textBF{.223}&.224&\textBF{.223}&		\textBF{.877}&.898&.898&.898&			\textBF{.826}&.905&.908&.905&		1.00&\textBF{.991}&\textBF{.991}&\textBF{.991}\\
8&\textBF{.289}&.304&.321&.309&				.506&.464&.449&\textBF{.424}&			.299&.307&.323&\textBF{.295}&		.917&.916&.922&\textBF{.915}\\
9&.563&.502&\textBF{.489}&.509&				\textBF{.766}&.820&.809&.853&			.802&\textBF{.799}&.808&.804&		.966&\textBF{.954}&.960&.960\\
10&\textBF{.474}&.593&.597&.576&			.723&\textBF{.715}&.783&.748&			\textBF{.790}&.815&.819&.811&		\textBF{.918}&.982&.985&.974\\
11&.089&\textBF{.054}&\textBF{.054}&\textBF{.054}&	\textBF{.720}&.749&.749&.749&			.984&\textBF{.976}&\textBF{.976}&\textBF{.976}&		1.00&\textBF{.988}&\textBF{.988}&\textBF{.988}\\
12&\textBF{.427}&.420&.434&.441&			.800&.811&\textBF{.736}&.763&			.720&.719&\textBF{.711}&.714&		1.00&1.00&1.00&\textBF{.995}\\
13&.139&\textBF{.135}&.139&\textBF{.135}&		.994&.994&\textBF{.992}&.994&			.983&\textBF{.982}&.983&\textBF{.982}&		.710&.710&.710&.710\\
14&\textBF{.409}&.450&.488&.451&			.839&.838&\textBF{.822}&.838&			\textBF{.730}&.740&.767&.741&		1.00&1.00&1.00&1.00\\
15&\textBF{.478}&.381&.399&.382&		.	.873&\textBF{.885}&.886&\textBF{.885}&		\textBF{.790}&.804&.818&.804&		1.00&1.00&1.00&1.00\\
16&.479&\textBF{.400}&.422&.401&			\textBF{.873}&.883&.891&.883&			\textBF{.790}&.807&.830&.807&		1.00&1.00&1.00&1.00\\
17&.120&\textBF{.059}&\textBF{.059}&\textBF{.059}&	\textBF{.607}&.610&.610&.610&			.939&.938&\textBF{.937}&.938&		1.00&.995&\textBF{.992}&.995\\
18&.351&.354&\textBF{.343}&.353&			.473&.451&.452&\textBF{.447}&			.479&.467&\textBF{.460}&.463&		.945&.949&\textBF{.937}&.947\\
19&\textBF{.404}&.408&.519&.407&			.816&.812&\textBF{.801}&.811&			\textBF{.727}&\textBF{.727}&.768&\textBF{.727}&		1.00&1.00&1.00&1.00\\
20&.171&.163&.174&\textBF{.149}&			.979&.980&\textBF{.974}&.975&			\textBF{.938}&.943&.941&.942&		.808&.789&.801&\textBF{.787}\\
21&.151&\textBF{.105}&\textBF{.105}&\textBF{.105}&	\textBF{.981}&.996&.996&.996&			\textBF{.923}&.961&.961&.961&		.897&\textBF{.845}&\textBF{.845}&\textBF{.845}\\
22&.323&\textBF{.315}&\textBF{.315}&.316&		.467&\textBF{.454}&.455&\textBF{.454}&		.488&.478&\textBF{.477}&.478&		.984&\textBF{.957}&.958&.958\\
23&.374&.351&\textBF{.343}&.353&			.577&\textBF{.564}&.568&.565&			.583&.572&\textBF{.570}&.573&		.999&.999&.999&.999\\
24&.329&\textBF{.312}&.333&.329&			\textBF{.753}&.820&.766&.805&			\textBF{.585}&.595&.591&.592&		.988&\textBF{.985}&.987&.988\\
25&\textBF{.059}&.081&.105&.081&			.868&\textBF{.866}&.870&\textBF{.866}&		.992&.986&\textBF{.982}&.986&		.996&.991&\textBF{.990}&.991\\
26&.407&\textBF{.322}&.327&\textBF{.322}&		\textBF{.715}&.793&.784&.793&			.678&.667&\textBF{.673}&.668&		1.00&\textBF{.979}&.982&\textBF{.979}\\
27&.399&\textBF{.318}&.325&.319&			\textBF{.723}&.790&.780&.790&			.672&\textBF{.667}&.674&\textBF{.667}&		1.00&\textBF{.972}&.976&\textBF{.972}\\
28&.372&.373&\textBF{.365}&.376&			\textBF{.550}&.608&.561&.595&			\textBF{.439}&.446&.435&.442&		.999&.999&.998&.999\\
29&.296&.307&\textBF{.247}&.328&			.587&.622&\textBF{.573}&.623&			.376&.384&\textBF{.357}&.395&		.988&.994&\textBF{.933}&.995\\
\hline
Rnk. & 3.000 & \textBF{1.966} & 2.603 & 2.431 	& \textBF{2.276} & 2.793 & 2.397 & 2.534	& 2.517 & 2.483 & 2.672 & \textBF{2.328}	& 3.138 & 2.276 & 2.379 & \textBF{2.207} \\ 
\hline
\end{tabular}

}
\end{sidewaystable*}
}

\subsection{Dataset properties}
\label{subsect:setCorr}

Correlation coefficients between classification quality measures and chosen set-specific characteristics were provided in tables~\ref{table:COR-Hamm} and~\ref{table:COR-microF1}. Among the investigated set-specific properties, our focus is put on label-density (LD), average imbalance ratio (avIR) and average scumble (AVsc). We are especially interested in those properties because they are strongly related to such undesirable phenomena as label imbalance, concurrence between rare and frequent labels.  

The first thing that should be noticed is that for the Hamming loss and zero-one loss there are significant differences in the correlation coefficients between the reference algorithm and proposed methods. The results show that the modified methods are less prone to imbalanced label distribution than the original LPW ensemble.

An interesting, though somewhat contrary to the expectations, result is that the FCM-W algorithm demonstrates a stronger correlation to average imbalance ratio than the remaining proposed approaches. This observation may suggest that the proposed weighting scheme leads the confusion matrix to overfit. Taking this into consideration, the application of FCM-O correction seems to be the best choice because it achieves the lowest ranks in terms of zero-one loss and micro-averaged $F_1$ loss. What is more important it tends to be less affected by class imbalance. 

%%Hamming
{
\begin{table*}
\centering
\tbl{Spearman correlation coefficient between the given set properties and the values of Hamming loss and macro $F_1$ loss respectively. Algorithms are numbered according to Section~\ref{sect:ExpSet}. Names of set-specific properties are abbreviated as in Table~\ref{table:Dataset_summ}. Negative values indicate that the loss decreases (the quality rises) when the corresponding property indicator increases. To provide a reference point to the results related to classification quality the table also shows average ranks achieved by classifiers. \label{table:COR-Hamm}}{
\begin{tabular}{@{}r|rrrr||}
  \hline
  &\multicolumn{4}{c||}{Hamming}\\
  \hline
 & 1 & 2 & 3 & 4 \\ 
  \hline
N & .125 & .065 & .067 & .067 \\ 
  d & -.011 & -.161 & -.153 & -.187 \\ 
  L & -.144 & -.372 & -.382 & -.385 \\ 
  LC & -.252 & -.360 & -.330 & -.329 \\ 
  LD & -.188 & -.049 & -.021 & -.025 \\ 
  avIR & .235 & .076 & .092 & .078 \\ 
  AVsc & .213 & .017 & .038 & .028 \\ 
  \hline
Rnk. & 3.000 & 1.966 & 2.603 & 2.431\\
   \hline
\end{tabular}%
\begin{tabular}{r|rrrr@{}}
  \hline
  &\multicolumn{4}{c}{macro $F_1$}\\
  \hline
 & 1 & 2 & 3 & 4 \\ 
  \hline
N & .138 & .185 & .220 & .175 \\ 
  d & -.054 & -.049 & -.008 & -.056 \\ 
  L & .467 & .490 & .505 & .472 \\ 
  LC & .097 & .147 & .086 & .112 \\ 
  LD & -.370 & -.358 & -.424 & -.367 \\ 
  avIR & .192 & .180 & .200 & .172 \\ 
  AVsc & .026 & .048 & .009 & .020 \\ 
  \hline
Rnk. & 2.276 & 2.793 & 2.397 & 2.534\\
   \hline
   
\end{tabular}

}
\end{table*}
}

%%micro
{
\begin{table*}
\centering
\tbl{Spearman correlation coefficient between the given set properties and the values of micro $F_1$ loss and zero-one loss respectively. \label{table:COR-microF1}}{
\begin{tabular}{@{}r|rrrr||}
  \hline
  &\multicolumn{4}{c||}{micro $F_1$}\\
  \hline
 & 1 & 2 & 3 & 4 \\ 
  \hline
N & .011 & .061 & .067 & .061 \\ 
  d & .166 & .132 & .163 & .132 \\ 
  L & .523 & .536 & .555 & .536 \\ 
  LC & -.238 & -.181 & -.167 & -.181 \\ 
  LD & -.676 & -.647 & -.643 & -.647 \\ 
  avIR & .366 & .322 & .327 & .322 \\ 
  AVsc & -.081 & -.097 & -.086 & -.097 \\ 
  \hline
Rnk. & 2.517 & 2.483 & 2.672 & 2.328\\
   \hline
\end{tabular}%
\begin{tabular}{r|rrrr@{}}
  \hline
  &\multicolumn{4}{c}{zero-one}\\
  \hline
 & 1 & 2 & 3 & 4 \\ 
  \hline
N & .288 & .250 & .307 & .310 \\ 
  d & .616 & .342 & .420 & .377 \\ 
  L & .488 & .229 & .234 & .253 \\ 
  LC & -.152 & -.231 & -.317 & -.179 \\ 
  LD & -.545 & -.390 & -.476 & -.373 \\ 
  avIR & .675 & .485 & .538 & .487 \\ 
  AVsc & .516 & .330 & .316 & .366 \\
  \hline
Rnk. & 3.138 & 2.276 & 2.379 & 2.207\\ 
   \hline
\end{tabular}
}
\end{table*}
}

\section{Conclusion}
\label{sect:Conc}

In this paper, we addressed the issue of applying correction methods based on fuzzy confusion matrix to improve the outcome of binary classifiers that constitute an LPW ensemble. To provide a reliable correction procedure we investigated three different ways of building a fuzzy confusion matrix. The conducted experimental evaluation allow us to make following conclusions:
\begin{itemize}
 \item The application of basic FCM correction procedure is not recommended since it is strongly affected by the label-imbalance ratio of the data.  The predictions of the corrected classifiers are skewed towards the majority class. 
 \item If the classification system is focused on improving the classification ratio of rare labels, it is desired to employ FCM-W correction method. 
 \item In general, it is recommended to use FCM-O system because it is slightly better than FCM-W system. 
\end{itemize}

The results are so promising that prompted us to continue research related to the use of fuzzy-confusion-matrix-driven correction methods in the field of multi-label classification. Focal points of our future research  are:
\begin{itemize}
 \item Building of methods aimed at dealing with imbalanced label distribution.
 \item Providing competence measures tailored to the corrected classifiers.
 \item Investigating the influence of changing base classifier model.
\end{itemize}

\nonumsection{Acknowledgements}
The work was supported by the statutory funds of the Department of Systems and Computer Networks, Wroclaw University of Science and Technology.
Computational resources were provided by PL-Grid Infrastructure.
%\clearpage
 \bibliography{fcmpw}
 \end{multicols}
\end{document}